\documentclass{ws-jmrr}

\usepackage{graphicx}
\usepackage{cite}
\usepackage{microtype}
\usepackage{multirow}

\graphicspath{{figs/}}

\newcommand{\ignore}[1]{%
}

\begin{document}

\markboth{J. Wang, T. Wu, I. Iordachita, P. Kazanzides}{Calibration and Evaluation of a Motion Measurement System for PET Imaging Studies}

\title{Calibration and Evaluation of a Motion Measurement System \\ for PET Imaging Studies}

\author{Junxiang Wang$^{1}$, Ti Wu$^{2}$, Iulian Iordachita$^{1}$, Peter Kazanzides$^{2}$}

\address{$^1$Dept. of Mechanical Engineering, Johns Hopkins University, Baltimore, MD 21218, USA \\
{\tt\small Email: [jwang334, iordachita]@jhu.edu}}

\address{$^2$Dept. of Computer Science, Johns Hopkins University, Baltimore, MD 21218, USA \\
{\tt\small Email: pkaz@jhu.edu}}

\maketitle

\begin{abstract}
  Positron Emission Tomography (PET) enables functional imaging of deep brain structures,
  but the bulk and weight of current systems preclude their use during many natural human activities,
  such as locomotion.
  The proposed long-term solution is to construct a robotic system that can support an imaging system
  surrounding the subject's head, and then move the system to accommodate natural motion.
  This requires a system to measure the motion of the head with respect to the
  imaging ring, for use by both the robotic system and the image reconstruction software.
  We report here the design, calibration, and experimental evaluation of a parallel string encoder mechanism
  for sensing this motion. Our results indicate that with kinematic calibration, the measurement system can achieve
  accuracy within 0.5\,mm, especially for small motions.
\end{abstract}


\begin{multicols}{2}
  \section{Introduction}

  Positron Emission Tomography (PET) relies on the injection of a radioactive tracer, which is then
  preferentially absorbed by specific tissues (based on the choice of tracer). The absorbed tracer emits
  positrons that react with nearby electrons, creating a pair of annihilation (gamma) photons that travel
  in opposite directions and are detected by the PET imaging ring.
  Higher sensitivity can be achieved by placing the PET detectors as close as possible to the subject.
  To accomplish brain imaging during locomotion and other natural activities, a wearable
  imaging device, such as the Helmet PET \cite{Majewski2011}, would be the ideal approach. Under current technology,
  however, sufficient sensitivity for neuroscience research would require a PET detector weighing 10\,kg or more
  \cite{Majewski2020} (other estimates are 15-20\,kg).
  Therefore, we are alternatively exploring the possibility of active compensation for head motion, where the PET
  imaging ring is suspended over the subject's head by a robotic system while the subject walks on a treadmill or engages
  in similar activities, as illustrated in Fig.~\ref{fig:system-concept}.
  The robotic system is tasked with keeping the imaging ring approximately
  centered around the subject's head (coarse motion compensation), while the image reconstruction algorithm corrects
  residual motion (fine motion compensation).
  This is an example of human-robot interaction, where the robot must safely move in proximity of the human.

  Given the high safety requirement when moving a 15-20\,kg weight over
  a human head with a robot, we plan to utilize both mechanical and optical measurement systems, and possibly also
  inertial sensing, to attain redundant sensing of the relative motion between the subject's head and the PET imaging ring.
  This paper addresses the design, calibration, and evaluation of a mechanical sensing system,
  consisting of six string encoders connecting the PET detector to a safety helmet attached
  to the subject's head. This system will provide the primary measurement for controlling the robot (coarse
  motion compensation) according to our current plan, due to its robustness and high-frequency measurements
  (on the order of 1\,kHz). On the contrary, optical sensing can achieve high accuracy, but at a cost of lower-frequency feedback
  (on the order of 30-60\,Hz) and risk of failing to provide a measurement.
  The decision on which measurement system is used for the purpose of image reconstruction (fine motion compensation) has not been made,
  as it involves evaluating the accuracy of the measurement systems and meeting the
  requirement for the reconstruction process. According to the imaging scientists, motion measurements for this task should be accurate
  to within about 0.5\,mm. Since the maximum radius of a human head is about 100\,mm \cite{Rodrigues2015},
  the accuracy requirement $\epsilon$ (mm) can be expressed as:
  \begin{equation}
    \epsilon = \Delta t + 100 \tan(\Delta \theta) \leq 0.5
    \label{eq:accuracy}
  \end{equation}
  where $\Delta t$ is the translation error (mm) and $\Delta \theta$ is the rotation error.
  In practice, this requirement will be stricter than necessary, since targets of interest will likely be closer
  to the center of the head than the above assumed 100\,mm.
  Nevertheless, evaluating the mechanical sensing system against this requirement is one of the goals of this paper.

  \begin{figurehere}
    \centering
    \vspace{1.5mm}
    \includegraphics[width=0.9\linewidth]{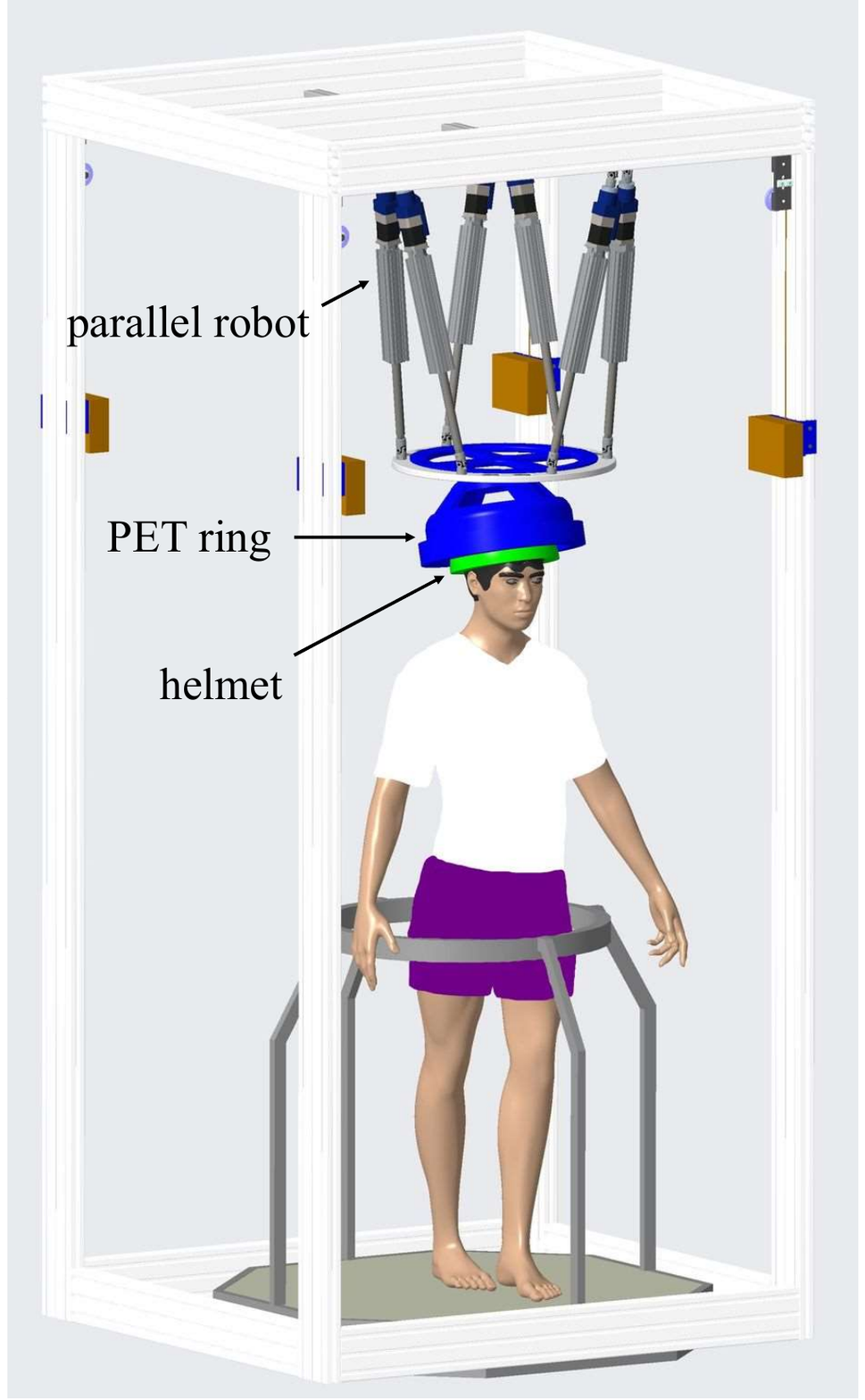}
    \caption{System concept: Parallel robot supports PET imaging ring around a subject's head, while
      the subject walks on a treadmill. The string encoder system measures the position of a helmet (green),
      worn by the subject, with respect to the imaging ring (blue).}
    \label{fig:system-concept}
  \end{figurehere}

  Motion correction is also relevant for conventional PET imaging, and several researchers have investigated
  different approaches for sensing this motion. Especially, different markerless approaches have been studied
  extensively in recent years. Olesen \cite{Olesen2011, Olesen2013} describes a system that incorporates a near
  infrared light emitting diode into a digital light processing projector to result in a surface scanner, tracking
  the head motion by creating a 3D point cloud on the head surface. This system is mounted directly on the PET
  imaging device, and while providing high accuracy of less than 0.3\,mm, adds considerable weight to the imaging ring.
  This setup also requires clearance between the patient's head and the imaging ring for the emission and
  processing of the near infrared light, which conflicts with our desire for a compact, highly-sensitive
  imaging system. Other methods rely more on image processing, thus reducing the required amount of computation
  compared to such a 3D surface imaging system. Kyme \cite{Kyme2018} and Anishchenko \cite{Anishchenko2015}
  utilized four cameras to record the patient's head, followed by detection of facial features from the video
  and determination of the 6 degree-of-freedom (DOF) movement of the head. However, this markerless approach
  could only achieve an accuracy of about 2\,mm.

  Chamberland \cite{Chamberland2011} uses positron emission fiducial markers to detect tumor movement
  for more accurate targeting in radiotherapy. This approach benefits from having motion be directly
  measurable in the images, which is ideal for the fine motion correction, but is not fast enough to
  control the robot for the coarse motion correction. Moreover, the fiducial marker would be visible in the
  reconstructed image and could interfere with perception of nearby brain structures.

  We previously determined the typical range of head motion, as well as velocity and acceleration, by analyzing motion
  capture and accelerometer data from subjects during overground and treadmill walking, respectively \cite{Liu2021}.
  We evaluated the ability of a simulated robot to compensate for the recorded head motion, without collision between 
  a helmet of dimension 263\,mm x 215\,mm and an imaging ring 300\,mm in diameter.
  Our results support the feasibility of such a compensation, given a robot capable of tracking typical head motions
  within about 100\,ms. This assertion will be experimentally verifiable with the measurement system that is the focus
  of this work.
  Section~\ref{sec:system} presents the measurement system design, implementation, and calibration;
  Section~\ref{sec:experiments} explains our means of evaluating the measurement system, including integration of a robot for
  providing ground-truth displacements. This is followed by calibration and accuracy results in Section~\ref{sec:results},
  and conclusions in Section~\ref{sec:conclusions}.
  The design of the measurement system and a preliminary accuracy evaluation (considering only translation measurement error
  under single-axis displacements)
  were reported in \cite{Wang2022}; this paper adds the kinematic calibration method (see Section~\ref{sec:calibration})
  and a more extensive accuracy evaluation, including translation and rotation errors under multi-axis displacements.

  \section{System}
  \label{sec:system}

  This section describes the design and construction of our mechanical measurement system using parallel string encoders,
  followed by a kinematic analysis and homing procedure (to determine initial string lengths).
  Then we discuss how the encoder attachment points and string lengths are adjusted using kinematic calibration.

  \subsection{Mechanical design}
  \label{sec:design}

  We designed a  parallel structure (essentially a passive parallel robot) consisting of 6 string encoders
  attached between the helmet worn by the subject and the imaging ring.
  For this system, forward kinematics converts the measured joint positions (string encoder lengths) to Cartesian pose.
  Since at least 6 string encoders are needed to provide 6 DOF, multiple configurations are possible.
  A Stewart platform structure was selected over other common cable-driven parallel robots (CDPRs) \cite{Lau2016}
  and parallel measuring structures \cite{Jeong1998} due to its compactness and prevalence \cite{Merlet2006},
  as well as its resistance against string interference (i.e., the strings should not come into contact
  with each other, the helmet, or the imaging device).

  \begin{figurehere}
    \centering
    \vspace{1.5mm}
    \includegraphics[width=0.98\linewidth]{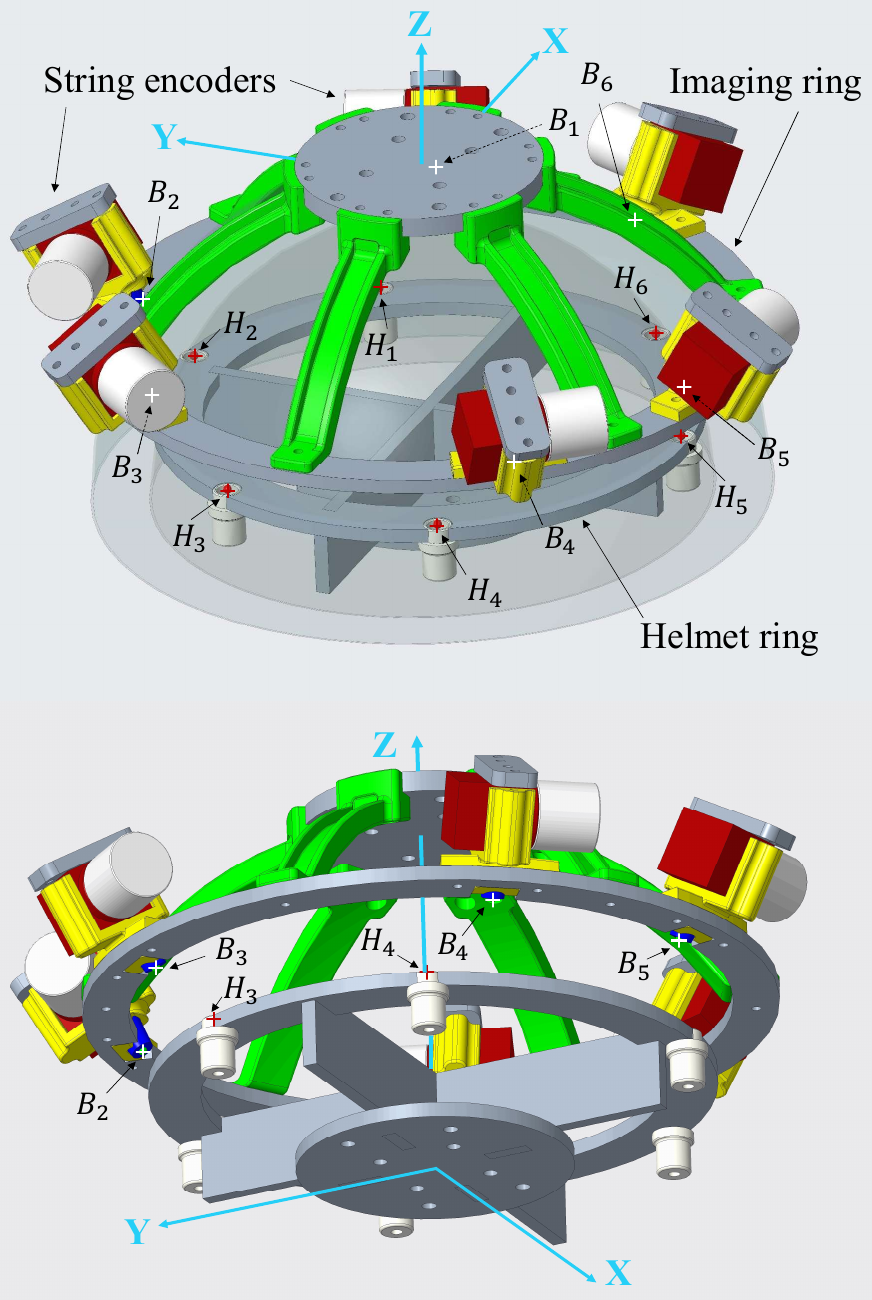}
    \caption{CAD model of string encoder system, with base (imaging ring) and helmet attachment points labeled}
    \label{fig:cad-model}
  \end{figurehere}

  Various Stewart platform designs were considered, evaluated based on motion sensitivity (resolution) and isotropic performance.
  The current design relies upon the approximation of the imaging ring as a sphere of 300\,mm diameter and the helmet/head as a sphere of
  250\,mm diameter. This approximation serves only for initial evaluation purposes, as the PET detector is expected to
  be cylindrical instead of spherical, and the helmet is expected to be elliptical with a major axis of about 263\,mm instead of circular.
  The prototype comprises two attachment rings---one represents the imaging ring, and the other represents the helmet.
  The placement of the rings within the respective spheres leads to the imaging ring having a diameter of 261.32\,mm and the helmet
  ring a diameter of 235.42\,mm.
  Fig.~\ref{fig:cad-model} depicts the string encoder system in two views of a CAD model generated with Creo.
  The locations of every pair of string attachment points are labeled in the top view angle, while a more direct view at the attachment points on the
  imaging ring is provided on the bottom. The orientation of the system is also shown in the figure: in the nominal configuration, the helmet sphere
  is concentric with the base (PET imaging ring) sphere, and the coordinate axes of the spheres are aligned as well.
  The designed string attachment points are given in Table~\ref{tab:attachment-points}, where $B_i$ and $H_i$ ($i=1\dots6$) are the base and
  helmet attachment points, respectively.
  \begin{tablehere}
    \centering
    \tbl{Designed Stewart platform attachment point coordinates (units: mm)}
    {
      \setlength\tabcolsep{4pt}
      \begin{tabular}{@{}crrrrrr@{}}
        \toprule
        \textbf{Encoder} & \multicolumn{3}{c}{$\boldsymbol{B_i}$} & \multicolumn{3}{c}{$\boldsymbol{H_i}$}                   \\
        \textbf{Number}  & \multicolumn{1}{c}{\textbf{X}} & \multicolumn{1}{c}{\textbf{Y}} & \multicolumn{1}{c}{\textbf{Z}} &
                            \multicolumn{1}{c}{\textbf{X}} & \multicolumn{1}{c}{\textbf{Y}} & \multicolumn{1}{c}{\textbf{Z}} \\
        \colrule
        1 &  121.39 &   48.35 & 73.67 &   96.99 &   66.70 & 42.06 \\
        2 &  -18.82 &  129.30 & 73.67 &    9.27 &  117.35 & 42.06 \\
        3 & -102.56 &   80.95 & 73.67 & -106.26 &   50.64 & 42.06 \\
        4 & -102.56 &  -80.95 & 73.67 & -106.26 &  -50.64 & 42.06 \\
        5 &  -18.82 & -129.30 & 73.67 &    9.27 & -117.35 & 42.06 \\
        6 &  121.39 &  -48.35 & 73.67 &   96.99 &  -66.70 & 42.06 \\
        \botrule
      \end{tabular}
      \label{tab:attachment-points}
    }
  \end{tablehere}

  \subsubsection{Construction}
  Each of the six string encoders (MPS-XXXS-200MM-P, Miran Industries, China) has a
  200\,mm measuring range at 60 counts/mm resolution.
  The string encoder bodies are secured onto the PET imaging ring and the mobile
  ends onto the helmet. The custom portions of the system comprise
  laser-cut acrylic plates and 3D-printed (Stratasys F170) ABS components. The
  geometry of the attachment rings and string connection points follows the Stewart
  platform presented above.

  The string encoders are incremental, with quadrature outputs (A and B channels) and
  an index pulse (I channel). A custom board was created to interface these signals to an FPGA board developed
  for the da Vinci Research Kit (dVRK) \cite{Kazanzides2014}. The FPGA firmware already included
  a four-channel encoder interface module, with modification applied to support a maximum of eight channels.
  The software for testing is implemented on a PC, which is connected to the FPGA board via a UDP socket.

  \subsection{String encoder kinematics}
  \label{sec:kinematics}

  The string encoder system adopts a Stewart platform structure; hence the standard kinematics model is applied,
  in which the inverse kinematics (from Cartesian pose to string lengths) is easily computed knowing
  the string attachment points on the base (PET imaging ring) and the moving platform (helmet).
  Let $X$ denote the Cartesian pose (transform) of the helmet with respect to the base, and $B_i$ and $H_i$
  ($i=1\dots 6$) denote the coordinates of the base and helmet attachment points, respectively. Then
  the computed string lengths, $\hat{L}_i$, are solved from the following inverse kinematics equation:
  \begin{equation}
    \label{eq:invkin}
    \hat{L}_i = \left | \left | X*H_i - B_i \right | \right |
  \end{equation}
  The forward kinematics involves higher complexity and requires numerical computation, as described in \cite{Harib2003}.
  Let $L_i$ denote the measured string lengths. Then the estimated Cartesian pose of the helmet with respect to the base,
  $\hat{X}$, is obtained from the following iterative computation:
  \begin{equation}
    \label{eq:fwdkin}
    \hat{X}_{k+1} = \hat{X}_k + J(\hat{L}_k) \left ( L_m - \hat{L}_k \right )
  \end{equation}
  where $k$ is the iteration counter, $J(\hat{L}_k)$ is the Jacobian and $\hat{L}_k$ is the inverse kinematics
  solution, eq. (\ref{eq:invkin}), corresponding to $\hat{X}_k$.
  The Cartesian pose here is a six-dimensional vector with the rotation component represented in Euler angles, following the intrinsic ZYX
  convention, i.e., with respect to the helmet coordinate system shown in Fig~\ref{fig:cad-model}.
  Subsequently in this paper, the rotation
  angle about the z-axis is referred to as roll, the angle about the y-axis as pitch, and the angle about the x-axis as yaw.
  The forward kinematics iteration is terminated when
  the combined error in string length, computed with $|| L_m - \hat{L}_k ||$, is below a specified threshold 
  (0.01\,mm in our implementation), hence signalling convergence. The second condition for termination is the iteration counter reaching a specified limit (50 in our
  implementation). Convergence is generally observed in three to four iterations.
  We would like to point out that for a parallel structure, the more practical approach to obtain the Jacobian is to compute the inverse Jacobian and then
  numerically solve for its pseudo-inverse.

  \subsubsection{Homing procedure}

  Since the string encoders only measure relative displacements, a homing procedure is required to set a reference absolute
  displacement, accomplished by utilizing the encoder index pulses transmitted on
  a separate channel from the regular quadrature signals. Through each encoder's 200\,mm range of travel, three index pulses
  are produced, evenly spaced apart by 4000 counts (one-third of the total measurement range).
  Each string encoder possesses a different position of the first index pulse, and the absolute displacements
  corresponding to these positions were recorded before the string encoders were installed.

  Based on this information, whenever the homing procedure needs to be performed, the first index pulse can be triggered by
  bringing the measuring ends as close to the encoder bodies as possible and subsequently
  captured by the FPGA firmware. Next, with calculation of the difference between these measured values and the recorded positions,
  offsets can be set in order to convert the relative displacements to absolute displacements.

  \subsection{Kinematic calibration}
  \label{sec:calibration}

  \subsubsection{Attachment point and string length adjustments}
  \label{sec:calibration-reason}

  The string encoder attachment points are influenced by several factors that lead
  to slight deviations from the kinematic model presented in Table~\ref{tab:attachment-points}. First of all, due to manufacturing inaccuracy,
  the real attachment points unavoidably differ from the exact values listed in Table~\ref{tab:attachment-points},
  and hence are in need of adjustment. On the other hand, we also performed an adjustment for the measured string
  lengths due to an observed measurement deficit. As depicted in Fig.~\ref{fig:string-hole}, the encoder has a circular
  guide channel for the string, and the center of that channel is the designed attachment point on the base,
  but the string instead lies against the edge of the channel in reality.
  This shift might be accounted for by adjusting the attachment point coordinates,
  along with the tuning needed due to manufacturing inaccuracy. However, during the helmet platform's motion,
  the string's position within the channel is also frequently shifted, especially when the motion is around the
  nominal position. This would hence require update of the attachment point coordinates during operation,
  complicating the approach. Instead, it was observed that because the guide channels are oriented at an angle, each
  string consistently remains in contact with the edge of the channel for a small range of motion around the nominal position. Thus, the actual
  reading reported by each encoder is less than the designed situation where the string exits from the center of
  the guide channel, and this difference always remains constant. Consequently, from kinematic calibration,
  we would like to determine for each string encoder this string length offset $\Delta L_i$, as well as updated
  coordinates for $B_i$ and $H_i$ that reflect manufacturing deviation from the designed values, hence a total of
  seven parameters.

  \begin{figurehere}
    \centering
    \vspace{1.5mm}
    \includegraphics[width=0.98\linewidth]{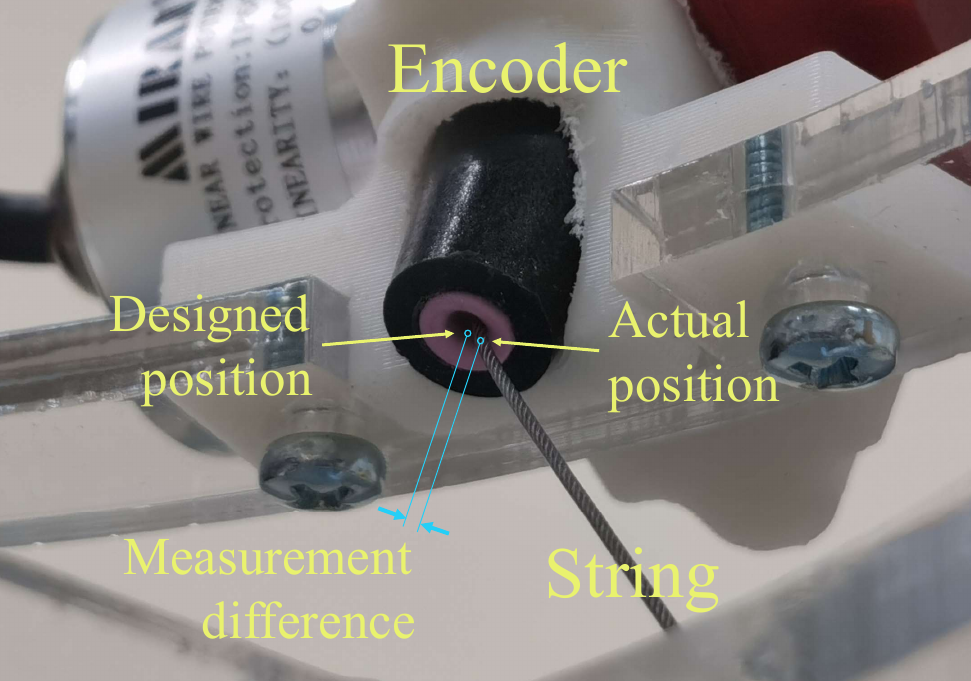}
    \caption{Deviation of the encoder string from the hole center}
    \label{fig:string-hole}
  \end{figurehere}

  \subsubsection{Parameter optimization}
  \label{sec:calibration-math}

  As described in Section \ref{sec:kinematics}, the forward kinematics for the Stewart platform possesses only a
  numerical solution, whereas the inverse kinematics has an analytical solution. Since the parameter optimization relies
  on taking derivatives, we chose to perform the minimization based on the inverse kinematics formulation, similar
  to approaches taken by other researchers \cite{Zhuang1995, Zhuang1998, Mayer1995}, by minimizing the string length
  measurement error in the Stewart platform's joint space. Moreover, performing calibration in inverse kinematics
  also provides the advantage of being able to solve the parameters of each leg independently, because the inverse
  kinematics equation (\ref{eq:invkin}) is specific to each leg and only depends on the platform pose in addition
  to the kinematic parameters of that leg. The following introduces the cost function of the minimization problem
  involved in the kinematic calibration, as well as the process of solving it using the Gauss-Newton algorithm.

  First, denote the collection of the seven parameters mentioned in Section \ref{sec:calibration-reason} as a vector:
  \begin{equation*}
    \boldsymbol{\theta}_i = \begin{bmatrix}
      B_i \\ H_i \\ \Delta L_i
    \end{bmatrix}
  \end{equation*}
  Next, suppose we have a ground-truth measurement of the Cartesian pose of the helmet with respect to the base,
  denoted as $X$. Then, the ground-truth string length is given by the inverse kinematics equation (\ref{eq:invkin})
  and consequently, if we denote the measured string length as $L_i$, then the error of this measurement is represented
  by the function:
  \begin{equation}
    f_i(\boldsymbol{\theta}_i) = L_i + \Delta L_i - \left|\left|X*H_i-B_i\right|\right|
  \end{equation}
  where $i=1\dots 6$ is the encoder index. If the helmet is moved to $n$ different poses, then for each encoder
  we have $n$ functions representing the measurement error. With $j=1\dots n$ as the pose index:
  \begin{equation}
    f_{i,j}(\boldsymbol{\theta}_i) = L_{i,j} + \Delta L_i - \left|\left|X_j*H_i-B_i\right|\right|
  \end{equation}
  Now, we are able to apply the Gauss-Newton algorithm to minimize the sum of squares of these $n$ functions for each
  encoder independently, that is, solve 6 minimization problems with the cost functions defined as:
  \begin{equation}
    C_i(\boldsymbol{\theta}_i) = \sum_{j=1}^n f_{i,j}(\boldsymbol{\theta}_i)^2
  \end{equation}

  The algorithm proceeds by first determining the Jacobian of the vector-valued function
  \begin{equation*}
    \boldsymbol{f}_i(\boldsymbol{\theta}_i) = \begin{bmatrix}
      f_{i,1}(\boldsymbol{\theta}_i) \\ \vdots \\ f_{i,n}(\boldsymbol{\theta}_i)
    \end{bmatrix}
  \end{equation*}
  If $R_j$ represents the rotation component of pose $X_j$, the Jacobian is given by:
  \begin{equation}
    \label{eq:jacobian}
    J_{\boldsymbol{f}_i}(\boldsymbol{\theta}_i) = \begin{bmatrix}
      \frac{\left(X_1*H_i-B_i\right)^\top}{\left|\left|X_1*H_i-B_i\right|\right|} & -\frac{\left(X_1*H_i-B_i\right)^\top R_1}{\left|\left|X_1*H_i-B_i\right|\right|} & 1      \\
      \vdots                                                                      & \vdots                                                                           & \vdots \\
      \frac{\left(X_n*H_i-B_i\right)^\top}{\left|\left|X_n*H_i-B_i\right|\right|} & -\frac{\left(X_n*H_i-B_i\right)^\top R_n}{\left|\left|X_n*H_i-B_i\right|\right|} & 1
    \end{bmatrix}
  \end{equation}
  Next, the Gauss-Newton algorithm states that the parameter update between iteration step $s$ and step $s+1$
  \begin{equation*}
    d\boldsymbol{\theta}_i = \boldsymbol{\theta}_i^{(s+1)} - \boldsymbol{\theta}_i^{(s)}
  \end{equation*}
  is obtained as the solution to the following linear system:
  \begin{equation}
    \label{eq:linear-system}
    J_{\boldsymbol{f}_i} \left ( \boldsymbol{\theta}_i^{(s)}\right )^\top J_{\boldsymbol{f}_i}\left (\boldsymbol{\theta}_i^{(s)} \right ) d\boldsymbol{\theta}_i=
    -J_{\boldsymbol{f}_i} \left ( \boldsymbol{\theta}_i^{(s)} \right )^\top \boldsymbol{f}_i \left ( \boldsymbol{\theta}_i^{(s)} \right )
  \end{equation}

  Starting from the initial guess of $\boldsymbol{\theta}_i^{(0)}$ with coordinates from Table~\ref{tab:attachment-points}
  and $\Delta L_i=0$, the iteration is repeated until the cost function between the previous and current sets of parameters
  changes by less than 0.01\%. The same collection of poses is used to calibrate parameters for all encoders.
  Details for obtaining the ground-truth poses are presented in the following section.

  \section{Experiments}
  \label{sec:experiments}

  This section presents our experimental setup that provides a ground-truth reference for the string encoder
  system using a robot, followed by the data collection process for kinematic calibration, as well as
  computation of measurement errors in accuracy evaluation. 

  \subsection{Experimental setup}

  \begin{figurehere}
    \centering
    \includegraphics[width=0.98\linewidth]{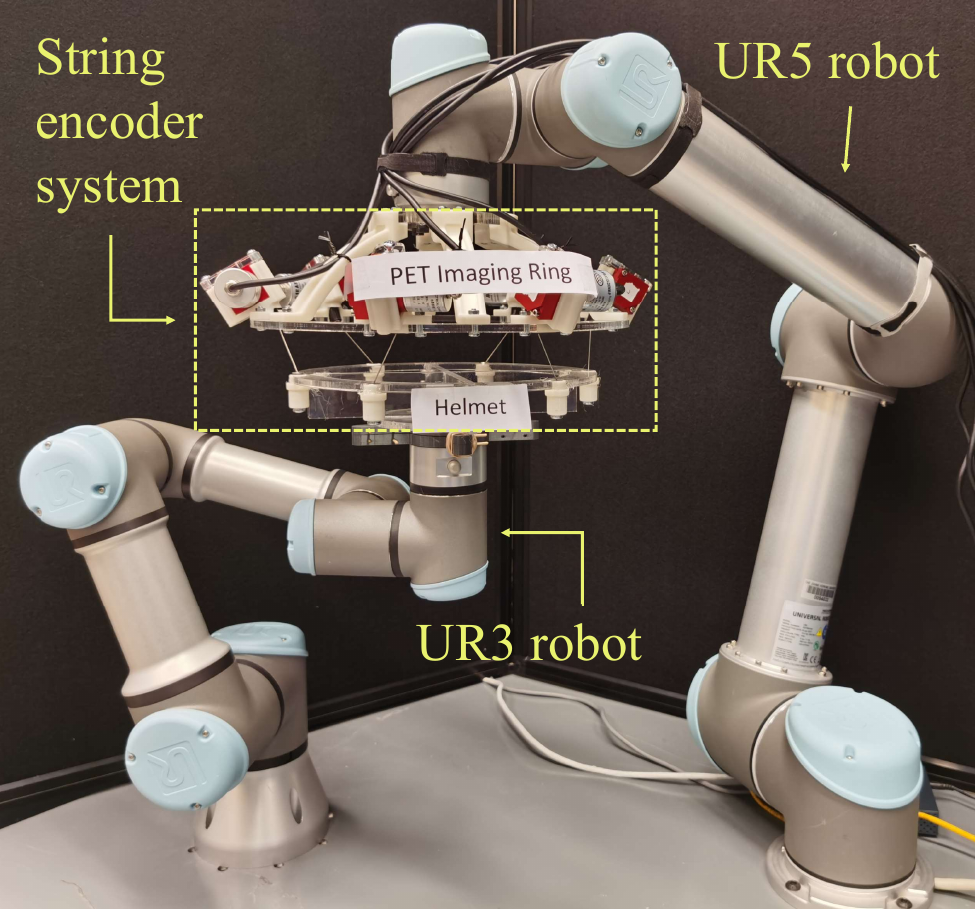}
    \caption{Experimental setup consisting of string encoder system and UR3 and UR5 robots}
    \label{fig:setup}
  \end{figurehere}

  To analyze the relative motion between the helmet (head) and the PET imaging system, in addition to the string
  encoder system described in Section~\ref{sec:system}, we employed an experimental setup consisting
  of a UR3 robot (Universal Robots, Odense, Denmark) to provide helmet ring movement and a UR5 robot for 
  supporting the PET imaging ring, as shown in Fig.~\ref{fig:setup}. In the experiments reported in this paper,
  the UR5 robot was not moved, whereas in future proof-of-concept work, its commanded motion will attempt to keep the
  (mock) PET imaging ring centered over the helmet based on the measured head position.
  In the longer term, the UR5 payload is insufficient to support the weight of an actual PET system
  and consequently a custom parallel robot, similar to the one shown in Fig.~\ref{fig:system-concept}, will be designed
  and will replace the UR5.

  Because the goal of the UR3 robot is to provide precise helmet motions, the measurement
  accuracy of the robot was first verified against a dial indicator (543-693B, Mitutoyo Corp.,
  Japan), for a $\pm$10\,mm range of motion. The robot motion is limited to this particular region since,
  as mentioned in Section \ref{sec:design}, the designed clearance between the helmet and the PET detector is only about 15\,mm.
  Since the UR3 end-effector is cylindrical, a part was designed and 3D-printed to provide
  a flat edge for ease of measurement. We interfaced to the UR3 robot via TCP, using its real-time script interface.

  Our previous work \cite{Wang2022} includes a figure illustrating this dial indicator setup, as well as the
  full result of the accuracy verification. We found that the difference between the commanded displacement
  to the robot and the measured displacement from the dial indicator generally stays less than 0.1\,mm, and
  the rough surface on the 3D-printed part could have partially contributed to this discrepancy.
  Nonetheless, it can be concluded that the accuracy of the UR3 robot is around 0.1\,mm and thus
  suffices to provide ground-truth helmet motions in the experiments conducted.

  \subsection{Transformation between robot and string encoder system}
  \label{sec:transform}

  Our calibration procedure requires ground-truth pose information, which can be provided by the UR3
  robot if its coordinate system is registered to the string encoder coordinate system.
  We accomplished this by considering that the transformation between the PET imaging ring (base) and the
  robot, $^BE_R$, is fixed because the UR5 robot remains stationary.
  After setting the robot tool center point (TCP) coincident with the center
  of the helmet, the measured pose outputted from the robot interface corresponds to the transformation between the robot and the
  helmet, $^HE_R$. Then, ${^BE_R}({^HE_R})^{-1} = {^BE_H}$, which is the transformation between the PET
  imaging ring and the helmet ring and the same transformation measured by the string encoder system.
  Since under the nominal position between the base and the helmet, this transformation is simply the
  identity, the UR3 robot only needs to query $^HE_R$ once at the nominal position to store it as $^BE_R$.
  Afterwards, the measurements from the UR3 robot can be converted to a 6 DOF pose in the string encoder
  coordinate system by computing ${^BE_R}({^HE_R})^{-1}$ as a transformation and extracting the translation
  and rotation components. For clarity and convenience, we denote this platform pose computed from robot
  measurements as $X^r$, and the platform pose computed with forward kinematics of the string encoder
  system as $X^e$.

  \subsection{Calibration procedure}
  \label{sec:calibration-procedure}

  The kinematic calibration is performed by moving the UR3 robot to various points within the workspace
  around the nominal position, recording the string lengths $L_{i,j}$, and using the robot-reported platform
  poses $X^r_j$ as the ground-truth poses in eq. (\ref{eq:jacobian}) to solve the minimization problem.
  The waypoints selected are all the points within a 15\,mm radius from the nominal position, with one at
  the center, and the rest equally spaced 5\,mm apart. The first set is collected with the rotation being the
  identity, and afterwards, the same waypoints are traversed five more times, each time with a different
  rotation, where each Euler angle is randomly sampled from a Normal distribution with zero mean and standard
  deviation of three degrees. This sums to an overall collection of 732 points, and as mentioned in Section
  \ref{sec:calibration}, this same collection can be used to calibrate the parameters for each leg
  independently. Since the linear system in eq. (\ref{eq:linear-system}) has more equations than unknowns, a
  least-squares solution is used to find the parameter update between iteration steps.

  \subsection{String encoder measurement accuracy evaluation}
  \label{sec:accuracy-methods}

  To perform the accuracy evaluation, we first designed a set of points that evenly explore the workspace of the
  helmet---up to 10\,mm or deg away from the nominal position. We accomplished this by forming 10 spherical shells,
  with the radii ranging from 1 to 10 in increments of 1. Each shell of radius $r$ contains 14 points evenly
  spaced apart---six along the coordinate axes (i.e. $[r,0,0]$, $[0,r,0]$, $[0,0,r]$ and their negatives) and one
  in the middle of each octant, with each coordinate equal to $\pm\sqrt{r/3}$ (for example, the coordinates in the
  first octant are given by $[\sqrt{r/3},\sqrt{r/3},\sqrt{r/3}]$). Summing the points in all ten
  shells results in a total of 140 points in $\mathbb{R}^3$. Interpreting these 140 points as Cartesian positions
  with mm as the unit yields the accuracy test set for translational motion. On the other hand, interpreting the
  same set in $\mathbb{R}^3$ as Euler ZYX rotation angles in degrees yields the test set for rotational motion.
  The robot was commanded to first execute purely translational motion, followed by purely rotational motion, with
  the test sets described above, in order to provide more information about the difference in accuracies associated
  with the two different types of motion. At each pose, the string lengths were recorded, adjusted by the offsets
  identified from kinematic calibration, and then the forward kinematics was applied under the calibrated attachment
  point coordinates, obtaining the 6 DOF Cartesian pose.

  The computed pose is denoted by $X^e=(t^e, R^e)$, where $t^e$ is the translational component and $R^e$ the rotational
  component, and similarly the ground-truth pose is $X^r=(t^r, R^r)$. The translation error of the
  string encoder system's measurement is
  \begin{equation}
    \Delta t = \left|\left|t^e\right|\right| - \left|\left|t^r\right|\right|
  \end{equation}
  and the rotation error is
  \begin{equation}
    \Delta \theta = \left|\left|\left(\ln R^e\right)^\vee\right|\right| - \left|\left|\left(\ln R^r\right)^\vee\right|\right|
  \end{equation}
  where the ``vee'' operator $^\vee$ denotes the conversion from a skew-symmetric matrix to the corresponding
  vector, and hence the rotation error represents the difference in the \emph{angle} value of the axis-angle
  representations of the measured and ground-truth rotations. Analogously, the translation error represents
  the difference in magnitudes between the measured and ground-truth translations, thus eliminating potential
  error caused by coordinate system registration.
  The accuracy of the string encoder measurement system is evaluated by calculating both the
  translation error and the rotation error at each position of the UR3. Moreover, for clarity, we use
  the subscript ``T'' to denote when the UR3 is performing a purely translational motion, and the subscript
  ``R'' for a purely rotational motion. For example, $\Delta t_R$ represents the translation measurement error
  when the UR3 is performing rotational motion only. We also compute the combined error, $\epsilon$, using
  equation (\ref{eq:accuracy}).

  \section{Results}
  \label{sec:results}

  \subsection{Calibration results}

  The kinematic calibration is carried out as mentioned previously, with the data collection process described in
  Section \ref{sec:calibration-procedure} and the optimization described in Section \ref{sec:calibration-math}.
  The optimization algorithm converges rapidly and meets the 0.01\% threshold after three to four iterations
  for all encoders.

  Table~\ref{tab:calibrated-params} shows the updated system parameters after calibration, including the base
  attachment point coordinates $B_i$, the helmet attachment points $H_i$, as well as the string length offset
  $dL_i$ for each encoder (hence each leg of the Stewart platform).
  \begin{tablehere}
    \centering
    \tbl{Calibrated Stewart platform kinematic parameters---attachment point coordinates and string length offsets (units: mm)}
    {
      \setlength\tabcolsep{2.8pt}
      \begin{tabular}{@{}crrrrrrr@{}}
        \toprule
        \textbf{Encoder} & \multicolumn{3}{c}{$\boldsymbol{B_i}$} & \multicolumn{3}{c}{$\boldsymbol{H_i}$} & \multirow{2}{*}{$\boldsymbol{dL_i}$} \\
        \textbf{Number}  & \multicolumn{1}{c}{\textbf{X}} & \multicolumn{1}{c}{\textbf{Y}} & \multicolumn{1}{c}{\textbf{Z}} &
                            \multicolumn{1}{c}{\textbf{X}} & \multicolumn{1}{c}{\textbf{Y}} & \multicolumn{1}{c}{\textbf{Z}} \\
        \colrule
        1 &  115.48 &   47.31 & 71.31 &   92.11 &   63.65 & 41.78 & 1.49 \\ 
        2 &  -16.43 &  128.18 & 70.46 &    9.31 &  114.63 & 42.16 & 0.65 \\ 
        3 & -100.45 &   79.93 & 70.69 & -103.85 &   48.06 & 41.49 & 5.43 \\ 
        4 & -100.82 &  -76.10 & 71.96 & -103.35 &  -48.84 & 43.14 & 4.05 \\ 
        5 &  -19.56 & -124.92 & 72.31 &    8.59 & -114.24 & 42.30 & 5.63 \\ 
        6 &  117.71 &  -46.15 & 75.75 &   94.38 &  -64.85 & 45.04 & 4.33 \\
        \botrule
      \end{tabular}
      \label{tab:calibrated-params}
    }
  \end{tablehere}

  The $dL_i$ values are all positive, suggesting that the raw measurements are short of the ideal measurements,
  which matches with the predicted measurement deficit as explained in Section \ref{sec:calibration-reason}.
  Furthermore, the attachment point coordinates between Tables \ref{tab:attachment-points} and
  \ref{tab:calibrated-params} each differ by less than 5\,mm, which is a reasonable representation of the
  potential errors in manufacturing and installation. The updated set of parameters is then used in the
  subsequent accuracy evaluation. 

  \subsection{String encoder system accuracy evaluation}

  The accuracy evaluation was performed as described in Section \ref{sec:accuracy-methods}. Table
  \ref{tab:accuracy-translation} shows the translation and rotation measurement errors when the ground-truth motion
  is translation only, accompanied by comparison between calibrated and uncalibrated kinematic parameters. Table
  \ref{tab:accuracy-rotation} shows the case with purely rotational ground-truth motion, and the data for both
  tables are represented in Fig.~\ref{fig:accuracy}. \newline

  \begin{figurehere}
    \centering
    \includegraphics[width=1.0\linewidth]{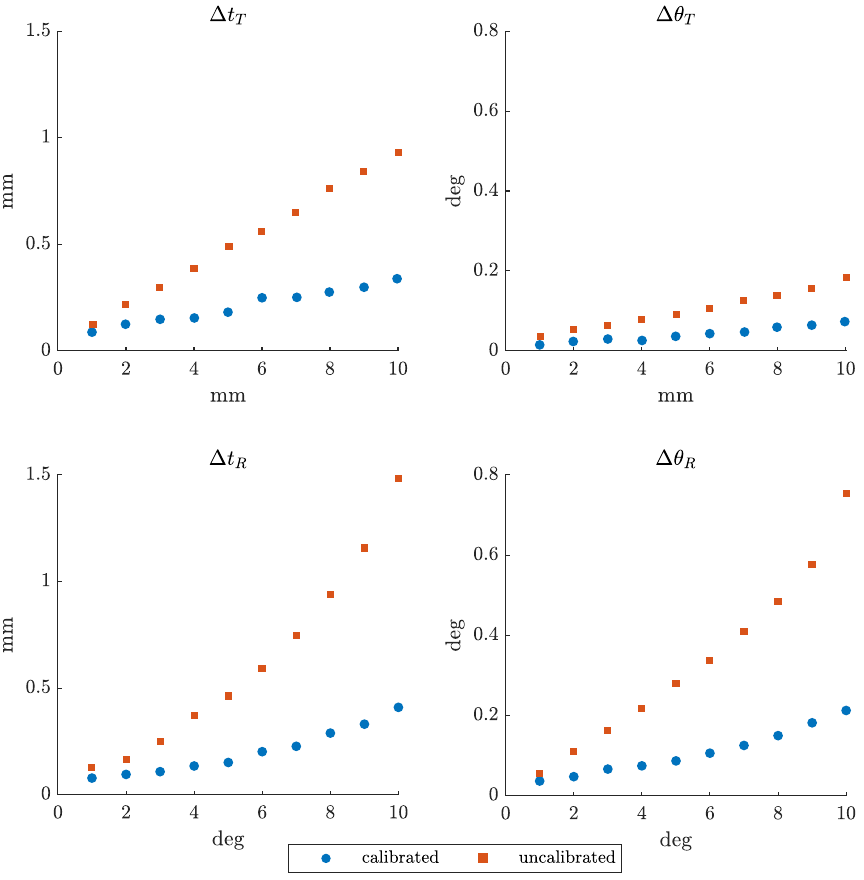}
    \caption{Accuracy evaluation result of the string encoder system, with comparison before and after calibration.
             Vertical axes represent translation error, $\Delta t$ (mm), and rotation error, $\Delta \theta$ (deg).
             Horizontal axes represent corresponding displacements (mm or deg).}
    \label{fig:accuracy}
  \end{figurehere}

  The results indicate that kinematic calibration reduced the overall RMS errors by 60-70\%. With the calibrated
  parameters, we have achieved translation RMS errors less than 0.3\,mm as well as rotation RMS
  errors less than 0.2\,deg for all motions, with more error associated with rotational motion than translational
  motion.
  The results also show that the combined error, $\epsilon$, is less than the 0.5\,mm threshold for translations
  up to 10\,mm and rotations up to 7\,deg. This indicates that the string encoder measurement system can meet
  the requirement for fine motion correction (during image reconstruction) assuming that the coarse motion
  correction (due to the robot moving the imaging ring) can keep the imaging ring within these translation and rotation limits.
  Further testing is required to determine whether these limits would need
  to be reduced to handle motions that include both translation and rotation. In addition, the current accuracy
  evaluation was performed under static conditions, and the error would likely be higher under dynamic conditions. \newline
  \begin{tablehere}
    \centering
    \tbl{RMS translation error, $\Delta t$ (mm), and rotation error, $\Delta \theta$ (deg),
      due to translational displacements (in mm) from the origin, with uncalibrated and calibrated parameters.
      $\epsilon$ is the combined error, eq. (\ref{eq:accuracy}).}
    {
      \setlength\tabcolsep{6pt}
      \begin{tabular}{@{}ccccccc@{}}
        \toprule
        \textbf{Displ.} & \multicolumn{3}{c}{\textbf{Uncalibrated}}       & \multicolumn{3}{c}{\textbf{Calibrated}}         \\
        \textbf{(mm)}   & $\Delta t_T$ & $\Delta \theta_T$ & $\epsilon_T$ & $\Delta t_T$ & $\Delta \theta_T$ & $\epsilon_T$ \\
        \colrule
         1 &  0.12 &  0.04 &  0.18 &  0.09 &  0.01 &  0.11 \\ 
         2 &  0.22 &  0.05 &  0.31 &  0.12 &  0.02 &  0.17 \\ 
         3 &  0.30 &  0.06 &  0.41 &  0.15 &  0.03 &  0.20 \\ 
         4 &  0.39 &  0.08 &  0.52 &  0.15 &  0.03 &  0.20 \\ 
         5 &  0.49 &  0.09 &  0.65 &  0.18 &  0.04 &  0.24 \\ 
         6 &  0.56 &  0.11 &  0.75 &  0.25 &  0.04 &  0.32 \\ 
         7 &  0.65 &  0.13 &  0.87 &  0.25 &  0.05 &  0.33 \\ 
         8 &  0.76 &  0.14 &  1.00 &  0.28 &  0.06 &  0.38 \\ 
         9 &  0.84 &  0.16 &  1.11 &  0.30 &  0.06 &  0.41 \\ 
        10 &  0.93 &  0.18 &  1.25 &  0.34 &  0.07 &  0.47 \\ 
        \colrule
        \textbf{Overall RMS} &  0.59 &  0.11 &  0.78 &  0.22 &  0.05 &  0.30 \\ 
        \botrule
      \end{tabular}
      \label{tab:accuracy-translation}
    }
  \end{tablehere}

  \section{Conclusions}
  \label{sec:conclusions}

  We developed a mechanical 6 DOF measurement system comprising six parallel string encoders, configured as a Stewart
  platform structure. The designed task of this system is to provide measurement of the motion of a helmet, worn by a human subject,
  relative to a PET imaging device supported by a robotic system. We performed kinematic calibration to improve the
  measurement accuracy of the system and conducted experiments with a robot providing
  precise helmet motions. The results indicate that the measurement accuracy can meet the 0.5\,mm requirement for fine
  motion correction during image reconstruction, assuming that the robotic system can keep the PET imaging ring ``near''
  the nominal (center) position. Based on our current results, ``near'' is defined as within 10\,mm and 7\,deg, but
  this definition may be changed in the future as more complex motions are evaluated under dynamic conditions.

  \begin{tablehere}
    \centering
    \tbl{RMS translation error, $\Delta t$ (mm), and rotation error, $\Delta \theta$ (deg),
         due to rotational displacements (in deg) from the origin, with uncalibrated and calibrated parameters.
         $\epsilon$ is the combined error, eq. (\ref{eq:accuracy}).}
    {
      \setlength\tabcolsep{6pt}
      \begin{tabular}{@{}ccccccc@{}}
        \toprule
        \textbf{Displ.} & \multicolumn{3}{c}{\textbf{Uncalibrated}}       & \multicolumn{3}{c}{\textbf{Calibrated}}         \\
        \textbf{(deg)}  & $\Delta t_R$ & $\Delta \theta_R$ & $\epsilon_R$ & $\Delta t_R$ & $\Delta \theta_R$ & $\epsilon_R$ \\
        \colrule
         1 &  0.13 &  0.05 &  0.22 &  0.08 &  0.04 &  0.14 \\ 
         2 &  0.17 &  0.11 &  0.36 &  0.09 &  0.05 &  0.18 \\ 
         3 &  0.25 &  0.16 &  0.53 &  0.11 &  0.07 &  0.22 \\ 
         4 &  0.37 &  0.22 &  0.75 &  0.13 &  0.07 &  0.26 \\ 
         5 &  0.46 &  0.28 &  0.95 &  0.15 &  0.09 &  0.30 \\ 
         6 &  0.59 &  0.34 &  1.18 &  0.20 &  0.11 &  0.39 \\ 
         7 &  0.74 &  0.41 &  1.46 &  0.23 &  0.12 &  0.44 \\ 
         8 &  0.94 &  0.49 &  1.79 &  0.29 &  0.15 &  0.55 \\ 
         9 &  1.16 &  0.58 &  2.16 &  0.33 &  0.18 &  0.65 \\ 
        10 &  1.49 &  0.75 &  2.80 &  0.41 &  0.21 &  0.78 \\ 
        \colrule
        \textbf{Overall RMS} &  0.76 &  0.40 &  1.45 &  0.23 &  0.12 &  0.44 \\
        \botrule
      \end{tabular}
      \label{tab:accuracy-rotation}
    }
  \end{tablehere}

  Our next stage of development will involve emulation of realistic human head motion with the UR3 robot, using the previously
  recorded motion data that our prior work analyzed \cite{Liu2021}. As the string encoder system collects measurement during the
  UR3 motion, we hope to have the UR5 robot and the PET imaging ring execute synchronized motion with the UR3 based
  on string encoder measurements. Eventually, since the UR5's 5\,kg payload falls short of the PET imaging ring mass of up to 20\,kg,
  it would be replaced by a custom robot, but currently it serves as a prototype for proof-of-concept experiments and verification of our
  motion measurement and compensation system.

  \section*{Acknowledgements}
  Stan Majewski initiated the investigation of robotic compensation for head motion during PET imaging.
  Yangzhe Liu participated in early design discussions of the measurement system.

  \bibliographystyle{ws-jmrr}
  \bibliography{references}
\end{multicols}
\end{document}